\documentclass[10pt,twocolumn,letterpaper]{article}

\usepackage[pagenumbers]{cvpr} 

\newcommand{\todo}[1]{{\color{black}#1}}

\definecolor{cvprblue}{rgb}{0.21,0.49,0.74}
\usepackage[pagebackref,breaklinks,colorlinks,allcolors=cvprblue]{hyperref}

\def\paperID{1309} 
\def\confName{CVPR}
\def\confYear{2025}

\usepackage{multirow}
\usepackage{amsmath}
\usepackage{graphicx}
\usepackage{ifthen}
\usepackage{color}
\usepackage[dvipsnames]{xcolor}
\definecolor{figure_roi_token}{RGB}{106,171,220}
\definecolor{figure_clip_token}{RGB}{178,211,164}

\usepackage{svg}
\usepackage{adjustbox}
\usepackage{array}
\usepackage{booktabs} 
\usepackage{subcaption} 
\usepackage{colortbl}
\usepackage{acronym}
\acrodef{roi}[RoI]{Region of Interest}

\newboolean{flag}
\setboolean{flag}{true} 
\setboolean{flag}{false} 

\newcommand{\specialtext}[1]{%
  \ifthenelse{\boolean{flag}}{%
    \textcolor{purple}{#1}%
  }{%
    #1%
  }%
}

\usepackage[utf8]{inputenc} 
\usepackage[T1]{fontenc}    
\usepackage{hyperref}       
\usepackage{url}            
\usepackage{booktabs}       
\usepackage{amsfonts}       
\usepackage{nicefrac}       
\usepackage{microtype}      
\usepackage{xcolor}         

\usepackage[capitalize]{cleveref}
\crefname{section}{Sec.}{Secs.}
\Crefname{section}{Section}{Sections}
\crefname{table}{Tab.}{Tabs.}
\Crefname{table}{Table}{Tables}
\crefname{equation}{Eq.}{Eqs.}
\crefname{figure}{Fig.}{Figs.}

\title{Context-Aware Aerial Object Detection: Leveraging Inter-Object \\and Background Relationships}

\author{
Botao Ren$^1$
\and
Botian Xu$^1$
\and
Xue Yang$^2$
\and
Yifan Pu$^1$
\and
Jingyi Wang$^1$
\and
Zhidong Deng$^1$\thanks{Corresponding author. Zhidong Deng is with Beijing National Research Center for Information Science and Technology (BNRist), Institute for Artificial Intelligence at Tsinghua University (THUAI), Department of Computer Science, State Key Laboratory of Intelligent Technology and Systems, Tsinghua University, Beijing 100084, China.}
\and
\ \\ $^1$Tsinghua University\\ $^2$OpenGVLab, Shanghai AI Laboratory
}

\begin{document}
\maketitle
\begin{abstract}
In most modern object detection pipelines, the detection proposals are processed independently given the feature map. Therefore, they overlook the underlying relationships between objects and the surrounding background, which could have provided additional context for accurate detection. Because aerial imagery is almost orthographic, the spatial relations in image space closely align with those in the physical world, and inter-object and object-background relationships become particularly significant. To address this oversight, we propose a framework that leverages the strengths of Transformer-based models and Contrastive Language-Image Pre-training (CLIP) features to capture such relationships. Specifically, Building on two-stage detectors, we treat Region of Interest (RoI) proposals as tokens, accompanied by CLIP Tokens obtained from multi-level image segments. These tokens are then passed through a Transformer encoder, where specific spatial and geometric relations are incorporated into the attention weights, which are adaptively modulated and regularized. Additionally, we introduce self-supervised constraints on CLIP Tokens to ensure consistency. Extensive experiments on three benchmark datasets demonstrate that our approach achieves consistent improvements, setting new state-of-the-art results with increases of 1.37 mAP$_{50}$ on DOTA-v1.0, 5.30 mAP$_{50}$ on DOTA-v1.5, 2.30 mAP$_{50}$ on DOTA-v2.0 and 3.23 mAP$_{50}$ on DIOR-R.
\end{abstract}    
\section{Introduction}
\label{sec:intro}

\begin{figure}[h]
    \centering
    \begin{subfigure}[b]{0.487\linewidth}
        \includegraphics[width=0.998\linewidth]{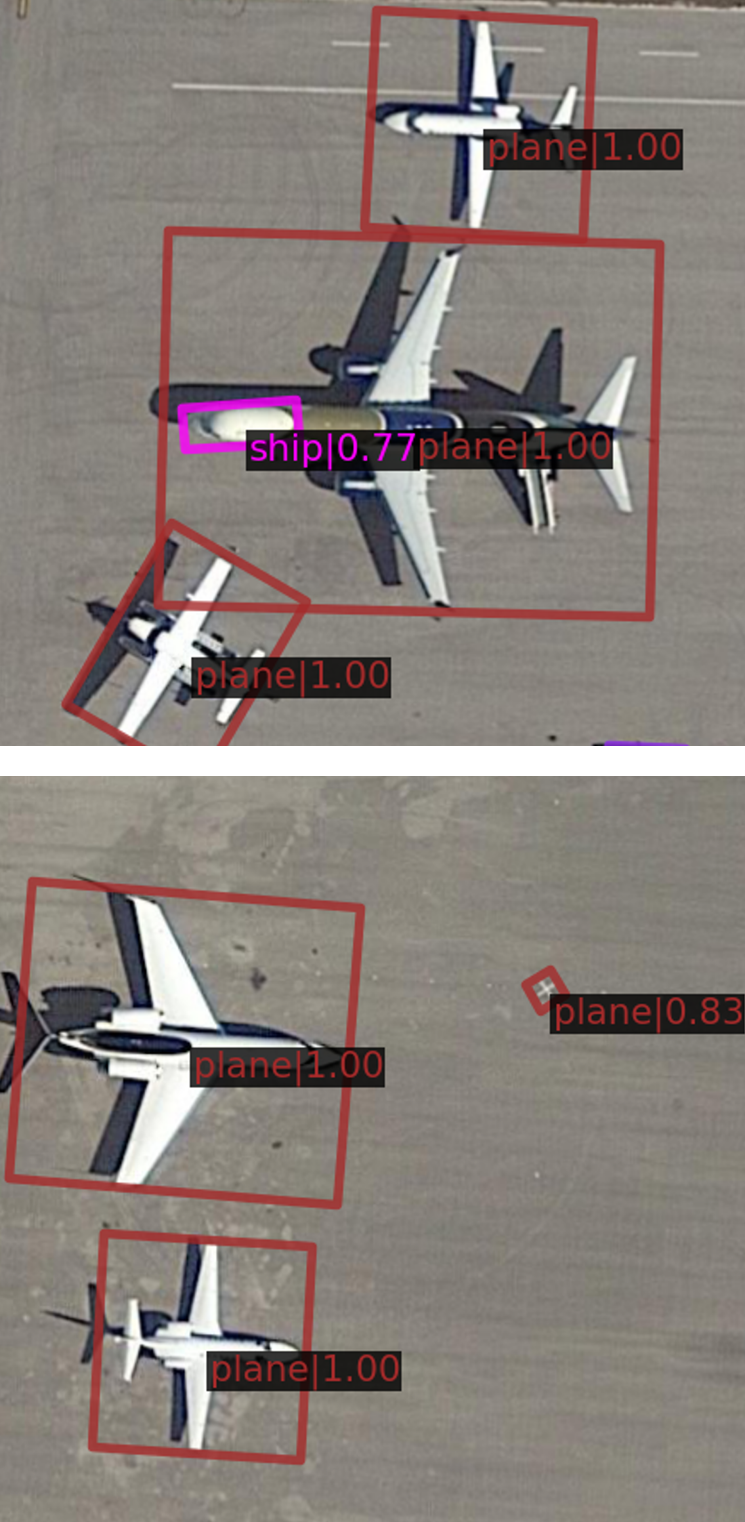}
        \caption{ReDet}
        \label{fig:sub1}
    \end{subfigure}
    \hfill
    \begin{subfigure}[b]{0.487\linewidth}
        \includegraphics[width=\linewidth]{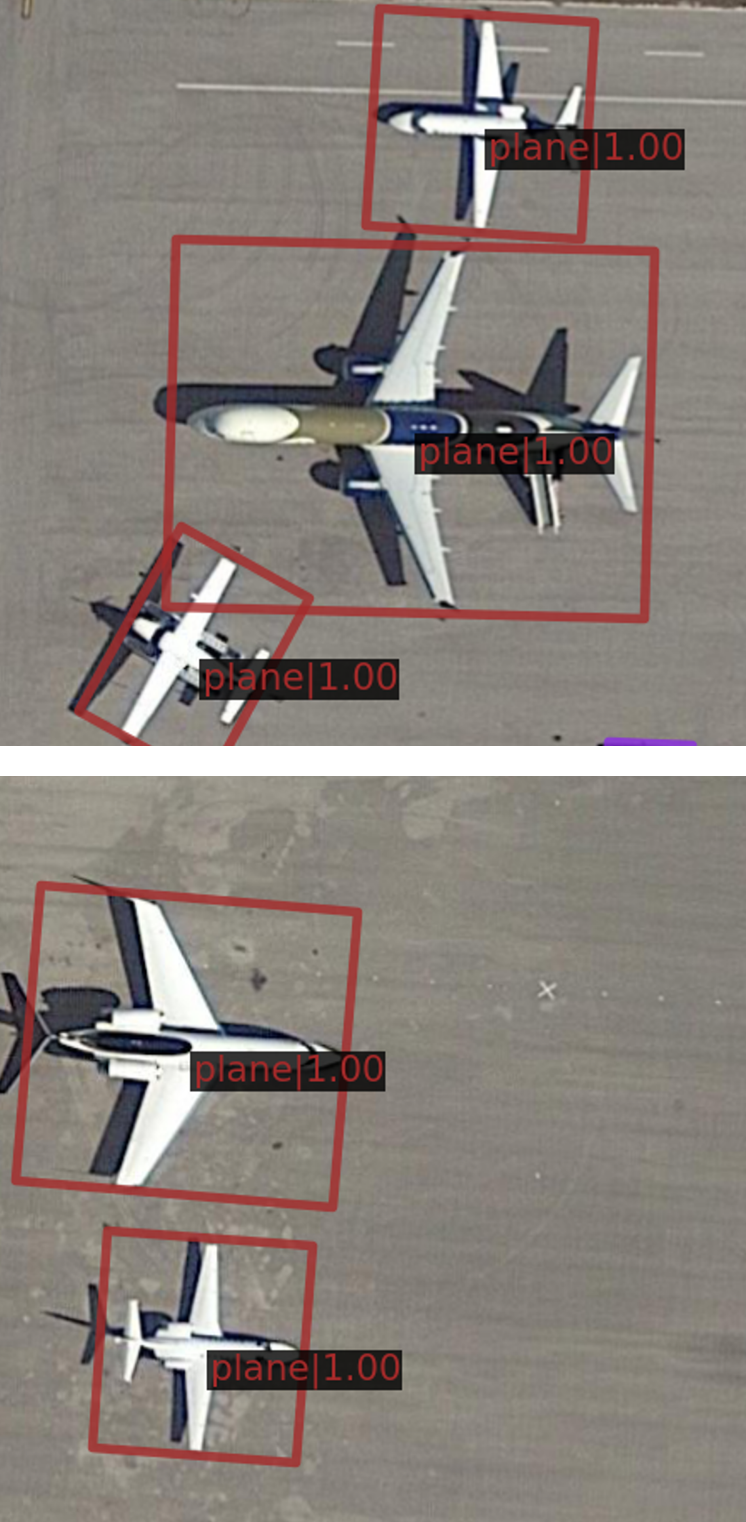}
        \caption{ReDet + Ours}
        \label{fig:sub2}
    \end{subfigure}
    \caption{Visualization of a motivating example. (a) Detections obtained by the ReDet \cite{han2021redet} have erroneous identifications: the upper left image shows a false detection of \textcolor{magenta}{\texttt{ship}} on top of an airplane; the bottom left image shows an incorrect \textcolor{BrickRed}{\texttt{airplane}} detection with unrealistic size. (b) Improved results obtained by our method. The false positives are effectively addressed. Highlighting the importance of considering inter-object relationship and background context in detection.
}
    \label{fig:example}
\end{figure}

Object detection has been one of the most studied problems in computer vision due to its great value in practical applications ranging from surveillance and autonomous driving to natural disaster management. The field has seen impressive advancements due to novel models and training techniques developed in the past few years \cite{liu2016ssd, redmon2018yolov3, carion2020detr}. Among various domains, object detection in aerial images stands out with characteristics and challenges different from those presented in natural images: objects are distributed with drastically varying scales, orientations, and spatial densities. 

To tackle this challenge, prior works proposed to improve the detection performance from different perspectives, achieving various degrees of success. Many efforts have focused on learning more appropriate features by exploiting the geometric properties, e.g., symmetry and rotational invariance, leading to novel architectures and data augmentation techniques \cite{han2021redet, xie2021oriented, yang2019scrdet, yang2021r3det, pu2023adaptive}. Others have developed metrics and objectives \cite{yang2021learning, yang2021rethinking, yang2022kfiou} that better capture the nuances of aerial object detection.

Nevertheless, despite these advancements, most present-day detection models classify and localize objects independently \cite{han2021redet, xie2021oriented, yang2019scrdet}, possibly due to the lack of an effective tool for modeling the co-presence of an arbitrary number of objects in an image. In other words, the spatial and semantic relationships among objects are not fully captured, often leading to false detections that overlook surrounding contextual dependencies and inter-object dynamics. As a motivating example, \cref{fig:example} illustrates the challenge of detecting each object instance based solely on its features, without considering these critical relationships. Aerial images, in particular, offer a unique setting where objects generally share the same plane, with little occlusion and perspective distortion, and therefore have stable inter-object and surrounding context relationships. Meanwhile, we posit that knowledge of an object's background context can provide useful information and therefore significantly improve detection. For instance, an area that appears as a green field might be presumed a playground; however, if adjacent to an airport, such an assumption would be reconsidered. Unfortunately, most datasets lack annotations for background information. The semantics of the background are complex and difficult to annotate due to the highly irregular spatial distribution.

In this paper, we propose a Transformer-based model on top of two-stage detectors to effectively capture and leverage the inter-object relationships and semantic background context. Concretely, we organize the Region of Interest (RoI) \cite{girshick2014rich, ren2015faster} feature maps proposed in the first stage and the independent detection results on them into embeddings. The embeddings are then fed into a transformer where the features of candidate detections interact and aggregate. However, the self-attention module in ordinary Transformers, which computes the pairwise attention weights as dot products of embeddings, does not capture the spatial and geometric relationship directly. To overcome this, we design and incorporate additional encodings and attention functions, weighing the mutual influence between objects according to distances. The attention functions are adaptive to the scales and densities of the object distribution, which is crucial for the model to generalize across different image scenarios. 

To further incorporate object-background relationships, we leverage CLIP \cite{radford2021learning}, a powerful multimodal model renowned for its cross-modal understanding capabilities, to integrate background information into detectors. Utilizing the image and text encoders of a pre-trained CLIP model, we divide the image into patches to be queried by pre-specified descriptions and then cast them as tokens alongside the RoI tokens.

\todo{Aerial images offer complex scenes with numerous objects on a single plane, where spatial and inter-object relationships are more explicit. The richer background context in such scenarios further supports the strengths of our approach.} We validate the effectiveness of our method through comprehensive experiments on DOTA-v1.0, DOTA-v1.5, DOTA-v2.0\cite{xia2018dota}, and DIOR-R \cite{cheng2022anchor}, achieving an improvement of 1.37, 5.30, 2.30 and 3.23 mAP$_{50}$ over the baseline.

Our main contribution can be summarized as follows:
\begin{itemize}
    \item We introduce a novel Transformer-based model that extends the capability of two-stage detectors, enabling the effective encapsulation and utilization of inter-object relationships in aerial image detection.
    \item We propose to use CLIP for integrating background context into the detection pipeline. We introduce multi-scale, hierarchical CLIP patches, generating CLIP Tokens and facilitating the flow of semantic information across different levels, thereby improving information fusion.
    \item Our model innovatively incorporates additional encodings and attention mechanisms that directly address spatial and geometric relationships, enhancing adaptability to the varying scales and densities in object distribution, a critical step forward for generalization in diverse aerial scenarios.
\end{itemize}

\section{Related Works}
\label{sec:formatting}

\subsection{Aerial Object Detection}
In the realm of aerial object detection, extensive research has been conducted to tackle the unique challenges posed by the diverse characteristics of aerial imagery. Numerous studies have explored both single-stage and two-stage methodologies. Notable two-stage methods include ReDet~\cite{han2021redet}, which focuses on handling scale, orientation, and aspect ratio variations, Oriented RCNN~\cite{xie2021oriented} that introduces improved representations of oriented bounding boxes, and SCRDet~\cite{yang2019scrdet} designed for addressing the challenges of dense clusters of small objects. Additionally, SASM\cite{hou2022shape}, Gliding vertex\cite{xu2020gliding}, and \ac{roi} Transformer\cite{ding2019learning} have contributed to the advancement of two-stage approaches. On the other hand, single-stage methods such as R\textsuperscript{3}Det\cite{yang2021r3det}, S\textsuperscript{2}ANet\cite{han2021align}, and DAL\cite{ming2021dynamic} have been developed, demonstrating the diversity of strategies employed in the pursuit of efficient aerial object detection. These methodologies often incorporate modifications to convolution layers, novel loss functions like GWD\cite{yang2021rethinking}, KLD\cite{yang2021learning}, and KFIoU\cite{yang2022kfiou}, as well as multi-scale training and testing strategies to enhance the robustness of object detection in aerial imagery. The evolving landscape of aerial object detection research reflects the ongoing efforts to address the complex challenges inherent in this field. In addition, ReDet\cite{han2021redet} and ARC\cite{pu2023adaptive} modified the convolution layers to explicitly cope with rotation.

\subsection{Capturing Inter-Object Relationships}
Common detection systems handle candidate object instances individually. They implicitly assume that the distribution of objects in an image is conditionally independent, which is generally false in reality. Transformer-based architecture \cite{vaswani2017attention, dosovitskiy2020ViT} has shown impressive capability in relational modeling across multiple domains. To address the oversight mentioned above, \cite{hu2018relation} introduced an object relation module that computes attention \cite{vaswani2017attention} weights from both geometric and appearance features of object proposal. The module is also responsible for learning duplicate removal in place of Non-Maximum Suppression (NMS), leading to an end-to-end detector. More recently, DETR \cite{carion2020end, he2022destr} formulates detection as a set-prediction problem and sets up object queries that interact with each other in a Transformer-decoder \cite{vaswani2017attention}. Its successors \cite{chen2022efficient, sun2021rethinking} improved the framework's efficiency by operating directly on features instead of object queries. Graph Neural Networks have also been explored as a powerful alternative in relation modeling for object detection and scene understanding. Typically, one constructs the graph with the objects being the nodes and the spatial relations as edges \cite{kim2022object, Zhao2023RGRNRG, Chen2021ObjectDU}. \cite{xu2019spatial} instead models region-to-region relations with learned edges. They also differ in how edges are obtained. In comparison to prior works, our method focuses on aerial images where the inter-object relationships are stable, with a more explicit design.

\subsection{CLIP Features}
The CLIP (Contrastive Language-Image Pretraining) \cite{radford2021learning} model is trained on a large corpus of image-text pairs and could be used to extract semantic information from images. CLIP has promoted the field of computer vision and is widely applicated in traditional vision tasks\cite{radford2021learning}. For example, the model and concepts of CLIP have been applied to a variety of other visual tasks, such as object detection, video action recognition and scene graph generation, showcasing its broad applicability and adaptability\cite{wu2023cora,zhao2022exploiting,wang2021actionclip,wang2023cross}. Its zero-shot classification capability allows it to accurately categorize images without additional fine-tuning on specific datasets. This feature is particularly useful in tasks and fields where labeled data is scarce. 
Building upon the CLIP model, RegionCLIP\cite{zhong2022regionclip} focuses on specific image regions for detailed semantic analysis. By focusing on distinct image regions, RegionCLIP can provide more precise and contextually relevant semantic information, which is critical for tasks that require the understanding of spatial relationships and localized features. 
In addition to CLIP and RegionCLIP, uniDetector\cite{wang2023detecting} shows the advancement of leveraging the rich information from both visual and language modalities.
As an object detection model, uniDetector combines the global context of transformers with the local feature extraction of CNNs, which is beneficial for visual tasks that need to handle objects of varying orientations and scales. Besides, the rich information from both visual and language modalities endows uniDetector with the ability to recognize open-vocabulary objects.
\section{Methodology}

\begin{figure*}[ht]
    \centering
    \includegraphics[width=0.95\textwidth]{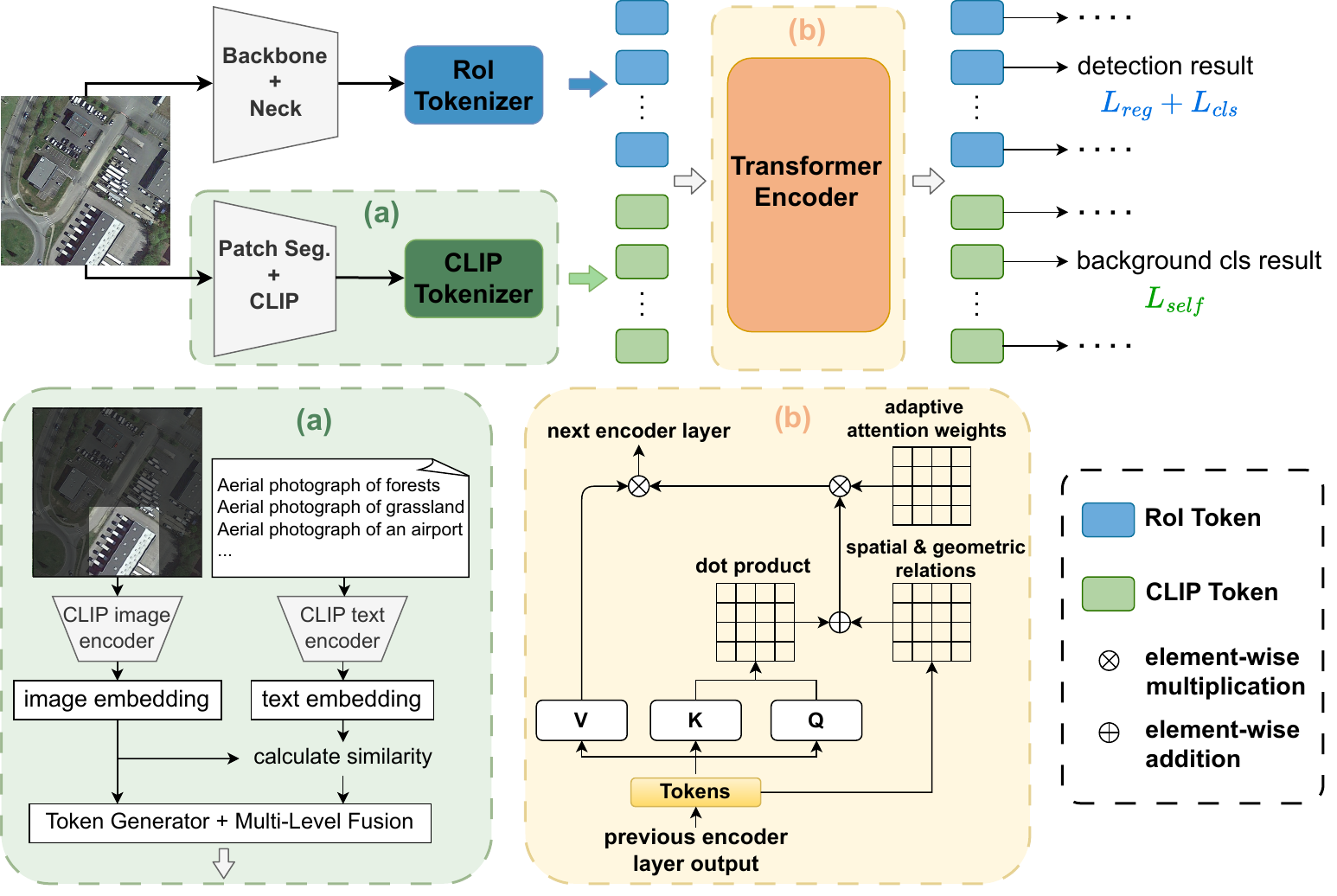}
    \caption{\textbf{Overview of our model.} The model utilizes a two-stage detection framework where features are converted into \textcolor{figure_roi_token}{\textbf{RoI tokens}}. Multi-scale patches generate \textcolor{figure_clip_token}{\textbf{CLIP tokens}} to capture background context, as shown in (a). Both tokens are processed by a Transformer encoder (b) with spatially aware attention, enhancing inter-object relationships based on distance and scale. Self-supervised constraints on CLIP tokens aid background classification, leading to refined detections with supervised signals \( L_{\text{reg}} + L_{\text{cls}} \) and \( L_{\text{self}} \).}
    \label{fig:overview}
\end{figure*}

We build our method on the two-stage object detection framework presented in \cite{han2021redet}. We start by following the original pipeline to obtain features and preliminary detections, which are then transformed into \textbf{RoI tokens}. Then we segment images into multi-scale patches and use CLIP to generate multi-level CLIP features, and then transform them into \textbf{CLIP tokens}. These tokens are input into the Transformer with additional encodings. To better leverage the Transformer, we introduce a novel attention function on top of the common scaled dot product and a set of spatial relations. It aims to reflect the degree of correlation between objects based on distances in the image, emphasizing neighboring detections while being aware of object scale and density. Moreover, we introduce self-supervised constraints on CLIP Tokens, providing additional supervised signals. Eventually, we perform another detection on the features given by the Transformer to obtain the final results. The overview of our model is shown in \cref{fig:overview}.

\subsection{RoI Tokens}
In a two-stage detector, the Region Proposal Network (RPN) proposes for each image $N$ RoIs from which we extract features $\{\mathbf{f}_i\}_{i=1}^N$. We apply the standard detection objective, namely classification and bounding box regression, on the features to obtain for each RoI a class label $c_i$ and a bounding box pose $(x_i, y_i, w_i, h_i, \alpha_i)$ representing the center coordinates, width, height and orientation, respectively.

Subsequently, we map $w_i, h_i$ through linear layers to high-dimensional embeddings $\mathbf{w}_i, \mathbf{h}_i$. They are then concatenated with the logits of the class distribution $\mathbf{c}_i$ to form the RoI Token:
\begin{equation}
    \text{Token}_i = (\mathbf{f}_i \oplus \mathbf{c}_i \oplus \mathbf{w}_i \oplus \mathbf{h}_i) + \text{pos}(x_i, y_i)
    \label{eq:roi_token}
\end{equation}
\specialtext{where $\oplus$ denotes concatenation} and the position encoding $\text{pos}(x_i, y_i)$ is computed as in \cite{dosovitskiy2020ViT} and added to enforce spatial information.

We will show in the experiments that, having \emph{two detection phases (preliminary and final)} is vital to the success of our model. This also distinguishes our work from prior works \cite{hu2018relation}.

\subsection{CLIP Tokens}


Besides RoI tokens, we additionally introduce CLIP Tokens to capture and articulate the multi-scale semantic context offered by the background. We use a CLIP model fine-tuned on the RSCID dataset fine-tuned on the RSCID dataset \cite{lu2017exploring}. We divide each image into patches of size $1$, $\frac{1}{2}$, $\frac{1}{4}$, and $\frac{1}{8}$, with strides 0, $\frac{1}{4}$, $\frac{1}{4}$, $\frac{1}{8}$, respectively. This results in 1, $5\times5$, $7\times7$, and $8\times8$ patches. The patches are resized and passed through the CLIP image encoder to get their image embeddings $\mathbf{f}_\text{image}\in \mathrm{R}^{139\times512}$. We also compute text embeddings using a set of pre-defined descriptions of the format "Aerial photograph of [\textbf{object}]", where \textbf{object} is selected from a range of natural and human landscapes such as forest, ocean, farmland, road, and airport. This results in $\mathbf{f}_\text{text}\in \mathrm{R}^{36\times512}$ from 36 descriptions. Incorporating the semantic information from text embeddings, CLIP tokens analogous to \cref{eq:roi_token} with $\mathbf{c}=\mathbf{f}_\text{image} \cdot \mathbf{f}_\text{text}^T$ and $\mathbf{w}$, $\mathbf{h}$ defined similarly according to patch sizes and locations. Moreover, to better combine information from patches of different sizes, we fuse them in a way similar to FPN. This part is depicted in \cref{fig:overview} {(a)}.

\subsection{Spatial and Geometric Relations}
\begin{table}[tb!]
\centering
\begin{tabular}{c|c|c}
\toprule
\textbf{Symbol}      & \textbf{Formula} & \textbf{Description} \\ \hline
$dx$       & $x_2-x_1$   & X-axis distance \\
$dy$       & $y_2-y_1$   & Y-axis distance \\
dist       & $\sqrt{dx^2 + dy^2}$ & Euclidean distance \\
$d\alpha$  & $ \alpha_2 - \alpha_1 $ & Angular difference \\
IoU       & $ intersect/union $ & Intersection over Union \\
\text{area} & $(w_1 h_1)/(w_2 h_2) $ & Relative area ratio \\ \bottomrule
\end{tabular}
\caption{Spatial and geometric relations considered in computing the attention weights between two RoIs.}
    \label{tab:relations}
\end{table}

Our method uses an encoder-only Transformer to capture the relationships between objects. However, the cosine distance self-attention computed between tokens in Transformers associates more closely to their semantic similarity but not spatial relations. Therefore, we introduce a series of $k$ relations $\{P^i\}_{i=1}^k$ accounting for the relative position and geometry between the preliminary detections as listed in \cref{tab:relations}. Similar to self-attention, each relation is computed in a pair-wise manner, i.e., $P^i\in \mathrm{R}^{N\times N}$. We concatenate them into a $N\times N\times k$ tensor and aggregate them to $N\times N\times 1$ by passing through a linear layer:
\begin{equation}
\label{eq:P}
    P = \text{Linear}(\text{stack}(P^1, \dots, P^k)).
\end{equation}

\subsection{Adaptive Attention Weights}
\label{sec:attention}

An inherent challenge in aerial images is that objects in a scene can vary drastically in size, orientation, and aspect ratio. Also, certain object types tend to cluster densely (like cars in parking lots) or align in specific patterns (such as parallel tennis courts). Thus the relationship between an object and others should be highly specific to the object instance and the contextual information around it. Based on this observation, we devise a novel scheme to adaptively adjust the attention weights with the following considerations. 

\paragraph{Spatial Distance, Scale, and Density.}

It is a natural intuition that the influence of one object on another relates to the distance between them and the relative scales (sizes) of the object. For example, closer objects are assumed to have a stronger correlation than distant pairs, and smaller objects tend to be more influenced by nearby objects, whereas larger objects need to capture the impact at longer ranges. The density around a proposed detection is another important factor. We assume that when there are fewer other RoIs around a detection (i.e., lower density), it should capture the influence of RoIs from further away. 
To qualitatively model these factors, we compute $\epsilon_i$ as:
\begin{equation}
\begin{aligned}
    A_{ij} = & \underbrace{\exp(-(\epsilon_i d_{ij})^2/\sigma^2)}_{\text{distance, scale and density}} \circ \ \underbrace{\mathbf{1}\{\text{IoU}_{ij}<\delta\}}_{\text{overlapping}}, \\
    \epsilon_i =& \frac{S}{\sqrt{w_i h_i}} \times \exp(\bar{\rho}_i),
    \label{eq:beta}
\end{aligned}
\end{equation}
where $d_{ij}=\sqrt{(x_i-x_j)^2+(y_i-y_j)^2}$ is the pair-wise distance, $\circ$ the element-wise product, and $\mathbf{1}\{\cdot\}$ the indicator function. $\epsilon_i$ is detailed next. $\sigma$ is a hyperparameter, and $S $ is a global (dataset-wide) scale factor determined by the input image.

To account for the density around an RoI, we first calculate for the $i$-th RoI:
\begin{equation}
    \rho_i = \sum_{i} w_i h_i \exp(-(\frac{S}{\sqrt{w_i h_i}} d_{ij})^2/\sigma^2),
\end{equation}
then image-wise normalize $\rho_i$ and map them into $(-1, 1)$:
\begin{equation}
    \bar{\rho}_i = \text{tanh}((\rho_i - \text{mean}(\rho)) / \text{std}(\rho)).
\end{equation}

\paragraph{RoI Overlapping.}

In addition to the aforementioned aspects, it is also necessary to mitigate the self-influence among multiple overlapping RoIs corresponding to the same object. Specifically, if we do not exclude these closely overlapping RoIs, their proximity to each other could lead to them being overly emphasized in the attention calculation while neglecting the interactions between RoIs of different objects. Therefore, we mask the attention weights to only consider RoIs with IoU below a certain threshold $\delta$.

The overall attention weights are calculated as 
\begin{equation}
    A \circ \text{softmax}(Q^\text{T} K + P)
    \label{eq:weights}
\end{equation}
where $P$ is the aggregated spatial and geometric relations computed in \cref{eq:P}.

\subsection{Loss Function}

We utilized a preliminary stage classification loss $L_{cls}^{pre}$, and final stage detection losses $L_{cls}$ and $L_{reg}$. Furthermore, to constrain CLIP tokens, we employed a self-supervised loss. The overall loss function is given by:
\begin{equation}
    L = L_{cls} + L_{reg} + \gamma L_{cls}^{pre} + \lambda L_{self}
\end{equation}
Where $L_{reg}$ is the standard bounding box regression loss, and $L_{cls}$ is the cross-entropy loss used in both stages. $L_{self}$ is the MSE loss between the background classification output $c_{clip}$ and its ground truth $c_{clip}^{GT}$. $\gamma$ and $\lambda$ are hyperparameters.
\section{Experiment}

\subsection{Dataset}


\begin{table*}[htbp] 
\centering
\small
\label{tab:dota-v1.5}
\resizebox{\linewidth}{!}{
\begin{tabular}{l|cccccccccccccccc|c}
\toprule
\textbf{Method}                               &  \textbf{PL}    & \textbf{BD}    & \textbf{BR}    & \textbf{GTF}   & \textbf{SV}    & \textbf{LV}    & \textbf{SH}    & \textbf{TC}    & \textbf{BC}    & \textbf{ST}    & \textbf{SBF}   & \textbf{RA}    & \textbf{HA}    & \textbf{SP}    & \textbf{HC}    & \textbf{CC}    & \textbf{mAP$_{50}$}   \\ \hline
RetinaNet-O\cite{lin2017focal}       &       71.43 & 77.64 & 42.12 & 64.65 & 44.53 & 56.79 & 73.31 & 90.84 & 76.02 & 59.96 & 46.95 & 69.24 & 59.65 & 64.52 & 48.06 & 0.83  & 59.16 \\
RF. R-CNN \cite{ren2015faster}        &         72.20 & 76.43 & 47.58 & 69.91 & 51.99 & 70.52 & 80.27 & 90.87  & 79.16 & 68.63 & 59.57 & 72.34 & 66.44 & 66.07 & 55.29 & 6.87 & 64.63 \\
Mask R-CNN\cite{he2017mask}          &       76.84 & 73.51 & 49.90 & 57.80 & 51.31 & 71.34 & 79.75 & 90.46 & 74.21 & 66.07 & 46.21 & 70.61 & 63.07 & 64.46 & 57.81 & 9.42  & 62.67 \\
HTC \cite{chen2019hybrid}            &       77.80 & 73.67 & 51.40 & 63.99 & 51.54 & 73.31 & 80.31 & 90.48 & 75.21 & 67.34 & 48.51 & 70.63 & 64.84 & 64.48 & 55.87 & 5.15  & 63.40 \\

\ac{roi}-Trans. \cite{ding2019learning}   &     72.27 & 81.95 & 54.47 & 70.02 & 52.49 & 76.31 & 81.03 & 90.90 & 84.19 & 69.12 & 62.85 & 72.73 & 68.67 & 65.89 & 57.09 & 7.12  & 66.69 \\

ReDet  \cite{han2021redet}           &     79.20 & 82.81 & 51.92 & 71.41 & 52.38 & 75.73 & 80.92 & 90.83 & 75.81 & 68.64 & 49.29 & 72.03 & 73.36 & 70.55 & 63.33 & 11.53 & 66.86 \\ 
FRED \cite{lee2024fred} & 79.60 & 81.44 & 52.60 & \textbf{72.57} & \textbf{58.07} & 74.82 & 86.12 & 90.81 & 82.13 & 74.84 & 53.37 & 72.93 & 69.51 & 69.91 & 54.82 & 19.27 & 68.30 \\
DCFL \cite{xu2023dynamic} & - & - & - & - & 57.31 & - & 86.60 & - & - & \textbf{76.55} & - & - & - & - & - & - & 70.24 \\

\rowcolor{gray!15} Ours (ReDet) & \textbf{80.17} & \textbf{83.71} & \textbf{54.28} & 70.31 & 52.80 & \textbf{77.42} & \textbf{88.46} & \textbf{90.84} & \textbf{86.02} & 75.04 & \textbf{68.19} & \textbf{73.09} & \textbf{76.94} & \textbf{74.51} & \textbf{73.86} & \textbf{28.86} & \textbf{72.16} \\ \bottomrule

\end{tabular}}
\caption{Results of each object class on the DOTA-v1.5 dataset. \todo{We use the standard 1x training schedule for fair comparisons.}}
\end{table*}
\begin{table}[]
\centering
\small
\resizebox{\linewidth}{!}{
\begin{tabular}{l|c|ccc}
\toprule
\textbf{Dataset}  &        \textbf{Method}            & \textbf{mAP$_{50}$}         &  \textbf{mAP$_{75}$}        & \textbf{mAP$_{50:95}$}           \\ \hline
\multirow{2}{*}{DOTA-v1.0} & ReDet  & 76.25        & 50.86        & 47.11        \\
   &      \cellcolor{gray!15}  Ours                    & \cellcolor{gray!15} 77.62 \textcolor{ForestGreen}{\footnotesize(+1.37)} & \cellcolor{gray!15} 52.18 \textcolor{ForestGreen}{\footnotesize(+1.32)} & \cellcolor{gray!15} 48.67 \textcolor{ForestGreen}{\footnotesize(+1.56)} \\ \hline
\multirow{2}{*}{HRSC2016}  & ReDet  & 90.46        & 89.46        & 70.41        \\
   &     \cellcolor{gray!15} Ours                    & \cellcolor{gray!15} 90.49 \textcolor{ForestGreen}{\footnotesize(+0.03)} & \cellcolor{gray!15} 89.67 \textcolor{ForestGreen}{\footnotesize(+0.21)} & \cellcolor{gray!15} 72.51 \textcolor{ForestGreen}{\footnotesize(+2.10)} \\ \bottomrule
\end{tabular}}
\caption{Results in COCO style on DOTA-v1.0 and HRSC2016.}
\label{tab:results2}
\end{table}
\begin{table}[]
\centering
\footnotesize
\resizebox{\linewidth}{!}{
\begin{tabular}{l|cccc}
\toprule
\textbf{Method}        & \textbf{DOTA-v1.0} & \textbf{DOTA-v1.5} & \textbf{DOTA-v2.0} & \textbf{DIOR-R} \\ \hline
Baseline      & 76.25     & 66.86     & 53.28     & 65.79  \\
\rowcolor{gray!15} Ours & 77.62 \textcolor{ForestGreen}{\footnotesize(+1.37)}     & 72.16 \textcolor{ForestGreen}{\footnotesize(+5.30)}     & 55.58 \textcolor{ForestGreen}{\footnotesize(+2.30)}     & 69.02 \textcolor{ForestGreen}{\footnotesize(+3.23)}  \\ \bottomrule
\end{tabular}}
\caption{\text{mAP}$_{50}$ results on DOTA-v1.0, DOTA-v1.5, DOTA-v2.0, and DIOR-R. For DOTA-v2.0, we use Oriented R-CNN as the baseline due to out-of-memory issues with the original ReDet. For the other datasets, ReDet is used as the baseline.}
\label{tab:result3}
\end{table}
\textbf{DOTA-v1.0} contains 2,806 images, with sizes ranging from $800\times800$ to $4000\times4000$ pixels. It includes 188,282 instances across 15 categories, annotated as: \texttt{Plane} (PL), \texttt{Baseball Diamond} (BD), \texttt{Bridge} (BR), \texttt{Ground Track Field} (GTF), \texttt{Small Vehicle} (SV), \texttt{Large Vehicle} (LV), \texttt{Ship} (SH), \texttt{Tennis Court} (TC), \texttt{Basketball Court} (BC), \texttt{Storage Tank} (ST), \texttt{Soccer Ball Field} (SBF), \texttt{Roundabout} (RA), \texttt{Harbor} (HA), \texttt{Swimming Pool} (SP), and \texttt{Helicopter} (HC). Following the common practice \cite{han2021redet}, we use both the training and validation sets for training and the test set for testing. We report mAP in PASCAL VOC2007 format and submit the testing result on the official dataset server.

\textbf{DOTA-v1.5} uses the same image set but with increased annotations. This version features 402,089 instances and introduces an additional category, \texttt{Container Crane} (CC), \specialtext{broadening the dataset's applicability. It also includes annotations for a greater number of small objects, some of which have areas smaller than 10 pixels, further enhancing the dataset's complexity.}

\textbf{DOTA-v2.0} further expands the datasets to 11,268 images and 1,793,658 instances, with two additional categories, \texttt{Airport} (AP) and \texttt{Helipad} (HP).

\textbf{DIOR-R} is a refined version of the original DIOR dataset, specifically re-annotated with rotated bounding boxes to enhance the detection of object orientation and shape in aerial images. It consists of 23,463 high-resolution images and 190,288 annotated instances, covering 20 diverse object categories, including vehicles, airplanes, ships, and more.

\textbf{HRSC2016} focuses on ship detection in aerial images, containing 1,061 images with a total of 2,976 instances. Image sizes in this dataset range from 300×300 to 1500×900 pixels. The dataset is divided into training, validation, and test sets with 436, 181, and 444 images, respectively.

\subsection{Implementation Details}
\label{sec:imple}
Our implementation is based on the MMRotate \cite{zhou2022mmrotate} library and adopts ReDet's framework and hyperparameter settings. We train our model for $12$ epochs using the AdamW \cite{loshchilov2018decoupled} optimizer with an initial learning rate of $1e^{-4}$, reduced to $1e^{-5}$ and $1e^{-6}$ at epochs 8 and 11. We also use a weight decay of 0.05. The experiments were conducted using two RTX 3090 GPUs.

The Transformer module consists of 6 encoder layers, similar to the ViT structure, and integrates sinusoidal two-dimensional absolute position encoding, hyperparameters $\sigma$ is set to 4. A dropout rate of 0.1 is employed during the training phase of the Transformer. In the loss function, we set $\gamma$ as 1, and $\lambda$ as 10.

\subsection{Comparison with Baselines}

First, we evaluate our model against the baselines on DOTA-v1.0, DOTA-v1.5, DOTA-v2.0, DIOR-R and HRSC2016 to demonstrate the efficacy of the proposed method. The results are shown in \cref{tab:results2}, and \cref{tab:result3}, respectively. 
These results demonstrate that our method consistently outperforms the baselines across different datasets. Notably, however, the improvement achieved in HRSC2016 is marginal compared to that on DOTA-v1.5 and DOTA-v2.0. This is possibly due to the number of instances in a single image being much fewer in HRSC (typically less than 4), thus there are limited opportunities to leverage the inter-object relationships. These findings suggest that our model's strengths are most pronounced in scenarios rich in object interactions and contextual dynamics, aligning with our design's focus on capturing and utilizing inter-object relationships.

\subsection{Ablation Study}


\paragraph{\textbf{Preliminary Detection Phase.}}
Compared to the standard detection pipeline, our model incorporates two detection heads - placed before and after the Transformer module. The output from the initial detection phase, dubbed \textbf{preliminary detection}, includes a classification result (parameterized as a softmax distribution) from the first head, which forms a component of the RoI token. We posit that knowing the class information with uncertainties would help with reasoning about the inter-object relationships. To empirically validate this hypothesis, we compared the performance of our model with and without training the first detection head. As \cref{tab:w_wo} shows, although solely incorporating the Transformer offers an improvement of mAP$_{50}$ to the baseline, omitting the preliminary detection leads to a notable decline in performance. This suggests that relying only on the Transformer for RoIs to interact lacks efficacy. In contrast, the explicit inclusion of preliminary classification data, despite its potential inaccuracies, enhances the model's ability to reason about semantical and contextual relationships. The results underscore the value of early classification cues in guiding the relational reasoning process within our proposed architecture.

\paragraph{\textbf{Spatial-Geometric Relations and Adaptive Attention Weights.}}
The different terms presented in \cref{tab:relations} characterize various aspects of the spatial and geometric relationships among objects (RoI Tokens) within an image. In this section, we aim to empirically evaluate the individual and collective contributions of these spatial and geometric relational terms to the overall performance of our detection model. As shown in \cref{tab:ab_relation}, IoU and rel. area contribute the most. Intuitively, they are particularly helpful when reasoning about the co-occurrence and spatial arrangement of objects. For example, IoU helps to disambiguate the overlapping, potentially duplicate or conflicting detections. Similarly, relative area aids in discerning the size relationship between objects. Consequently, our method can effectively solve the problem in the motivation example. See \cref{sec: quantitative} for details.

As mentioned in \cref{sec:attention}, making the attention weights adaptive to specific RoI Tokens is essential to cope with the diversity and complexities in a scene. We evaluate our density- and scale-aware attention weighting scheme which is designed to augment the scaled-dot-product self-attention and allow the model to dynamically adjust its focus based on the scale of objects and their surrounding density. Findings in \cref{tab:ab_relation} indicate that masking the influence of overlapping RoIs plays a crucial role. This observation aligns with our initial understanding that indiscriminately emphasizing neighboring RoIs, without considering overlap, could lead to skewed attention distributions and potentially impair the model’s ability to accurately discern between distinct objects.

\begin{table}[]
\centering
\resizebox{0.65\linewidth}{!}{
\begin{tabular}{cc|c}
\toprule
\multicolumn{2}{c|}{\textbf{Module}}                              &  \textbf{mAP$_{50}$} \\ \hline
\multirow{2}{*}{Pre cls supervision}  & w/o     & 69.86                               \\
                                      & w   & 71.74                               \\ \hline
\multirow{2}{*}{Self supervised loss} & w/o     & 71.89                               \\
                                      & w   & 72.16                               \\ \hline
\multirow{2}{*}{Multi-level fusion}   & w/o     & 71.17                               \\
                                      & w   & 72.16                               \\ \bottomrule
\end{tabular}}
\caption{Ablation of detection performance w/ or w/o Preliminary detection training, Self-supervised loss, Multi-level fusion.}
\label{tab:w_wo}

\end{table}

\begin{table}[]
\centering
\resizebox{\columnwidth}{!}{
    \begin{tabular}{cccc|c}
    \toprule
    \textbf{RoI Tokens} &  \textbf{CLIP Tokens} & \textbf{Relations} & \textbf{Ada. Weight} & \textbf{mAP$_{50}$} \\ \hline
                &            &           &           & 66.86 \\
    \checkmark  &            &           &           & 69.02 \\
    \checkmark  & \checkmark &           &           & 70.96\\
    \checkmark  & \checkmark &  \checkmark &         & 70.87 \\
    \checkmark  & \checkmark &            &\checkmark & 71.61 \\ 
    \rowcolor{gray!15} \checkmark  & \checkmark & \checkmark &\checkmark & \textbf{72.16} \\ \bottomrule
    \end{tabular}
}
\caption{Effect of the components in computing the attention weights via \cref{eq:weights}. Relations are calculated in \cref{eq:P}, and Adaptive weight is calculated in \cref{eq:beta}}
\label{tab:ab_attention}
\end{table}

\subsection{Self-supervised and Multi-level Fusion}

We show the additional improvements achieved by incorporating the self-supervised loss and multi-level fusion for CLIP tokens in \cref{tab:w_wo}. Self-supervised loss effectively regularizes CLIP tokens to prevent representation collapse. And multi-level fusion helps capture information at different spatial scales.

\begin{table}[]
\centering
\begin{tabular}{cccccc|c}
\toprule
$dx$        & $dy$        & $d\alpha$        & dist      & IoU       & area      & \textbf{mAP$_{50}$} \\ \hline
\checkmark & \checkmark &           &           &           &           & 70.79      \\
\checkmark & \checkmark & \checkmark & \checkmark &           &           & 71.03      \\
\checkmark & \checkmark & \checkmark & \checkmark & \checkmark &           & 71.46      \\
\rowcolor{gray!15} \checkmark & \checkmark & \checkmark & \checkmark & \checkmark & \checkmark & 72.16      \\ \bottomrule
\end{tabular}
\caption{Effect of the spatial and geometric relations.}
\label{tab:ab_relation}
\end{table}
\begin{table}[ht]

        \centering
        \begin{tabular}{c|c}
        \toprule
        \textbf{Attention Weights ($\beta$) }  & \textbf{mAP$_{50}$}   \\ \hline
        baseline (ReDet)    & 66.86 \\
        \specialtext{- scale ($\epsilon=\sqrt{S}=32$)}       & 70.49 \\
        \specialtext{- density ($\bar{\rho}_i=0$)} & 71.30 \\
        \rowcolor{gray!15} Ours        & \textbf{72.16} \\ \bottomrule
        \end{tabular}
        \caption{Effect of the factors in computing $\beta$.}
        \label{tab:ab_adaptive}
\end{table}
\section{Analysis}

To gain insights into how inter-object relationships have improved detection performance, we collect and analyze dataset-wise statistics and specific examples.

\begin{figure*}[ht]
    \centering
    \includegraphics[width=0.95\textwidth]{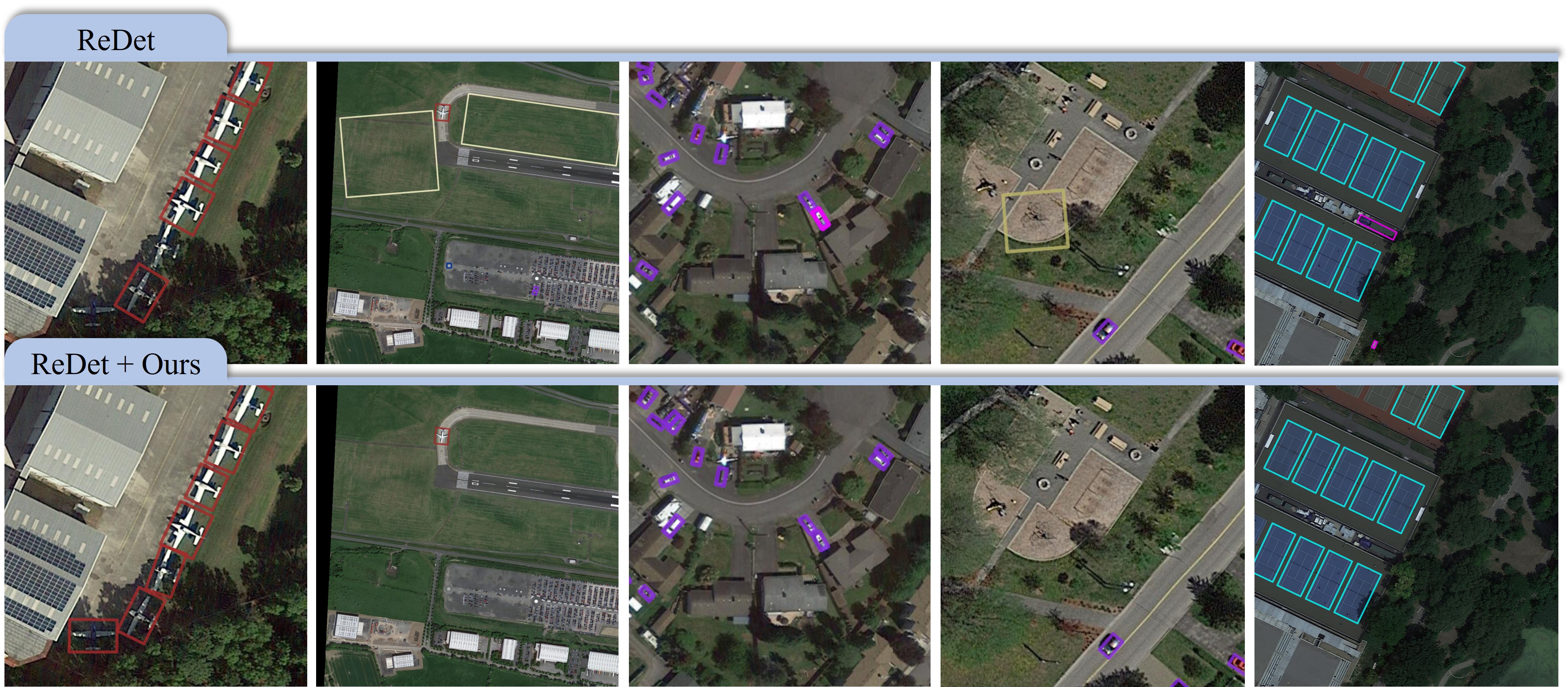}
    \caption{Qualitative comparison of our model and ReDet.}
    \label{fig:visualization}
\end{figure*}

\subsection{Evaluation Statistics}
\begin{figure}[ht]
    \centering
    \includegraphics[width=0.46\textwidth]{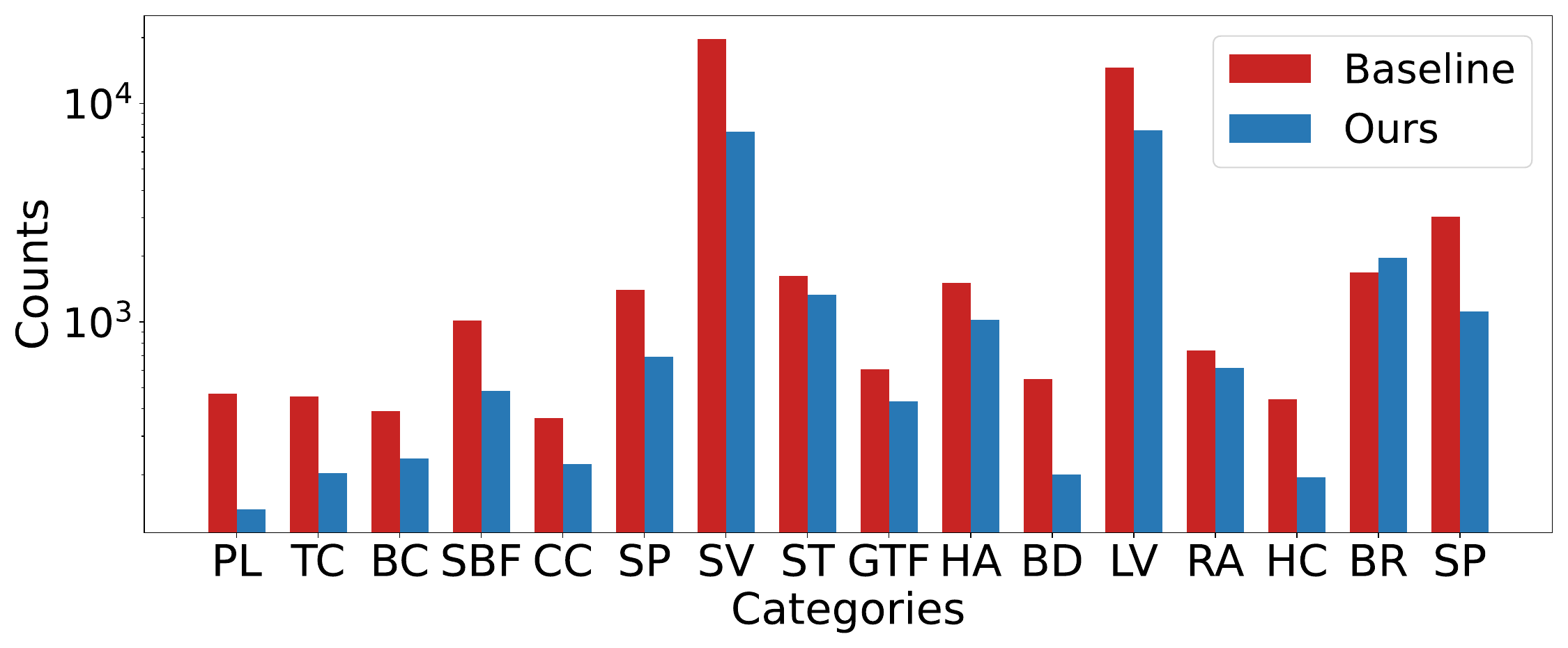}
    \caption{Count of outliers (in log-scale) for each category on the test dataset. }
    \label{fig:outliers}
\end{figure}

By examining the data we found that many false detections deviate far from the typical scales associated with their respective categories. To investigate this observation, we compute for each category the mean and standard deviation of object scale $\sqrt{w_i h_i}$ using detections with confidence $>0.9$ on the test set. We then identified outliers as those detections deviating from the mean by more than three times the standard deviation. This method provides a rough measure of the frequency of incorrect scale detections. As shown in \cref{fig:outliers}, the detections produced by our methods have substantially fewer outliers compared to the baseline. This result suggests that our model better maintains scale consistency across different object categories. This improvement is particularly vital in aerial image analysis, where scale variance is substantial and often indicative of the detection model's reliability and robustness.

\label{sec: quantitative}

Additionally, our visual analysis revealed a common misclassification of many land-based objects as \texttt{Ship}. To quantify this observation, we compute the average chamfer distance between certain categories $S_1$ and $S_2$ in an image:
\begin{equation}
\begin{split}
    d(S_1, S_2) &= \frac{1}{2} \left( \frac{1}{|S_1|} \sum_{i \in S_1} \min_{j \in S_2} \| \text{dist}_{ij} \|^2_2 \right) \\
    &\quad + \frac{1}{2} \left( \frac{1}{|S_2|} \sum_{j \in S_2} \min_{i \in S_1} \| \text{dist}_{ij} \|^2_2 \right)
\end{split}
\end{equation}

\begin{table}[tb!]
\centering
\label{tab:chamfer}
\begin{tabular}{l|ll}
\toprule
\textbf{Categories}            & \textbf{Baseline}  & \textbf{Ours}    \\ \hline
\tt Ship$\Leftrightarrow$Small Vehicle & 504.68  & 844.93 $\uparrow$  \\
\tt Ship$\Leftrightarrow$Plane         & 1000.48 & 1864.75 $\uparrow$ \\
\tt Harbor$\Leftrightarrow$Ship        & 266.54  & 238.62 $\downarrow$ \\
\bottomrule
\end{tabular}
\caption{Average Chamfer distance between detections of ship, small-vehicle, plane, and harbor.}
\label{tab:chamfer}
\end{table}

As Table \ref{tab:chamfer} shows, the results are in line with the logical expectation that \texttt{Ship} instances should be found in water, near \texttt{Harbor}, but distant from \texttt{Small Vehicle}. This finding underscores our model's effectiveness in accurately understanding the spatial arrangement of objects, further validating the benefits of our approach in handling complex aerial imagery.

\subsection{Limitations}

\begin{figure}[tb!]
    \centering
    \includegraphics[width=0.98\linewidth]{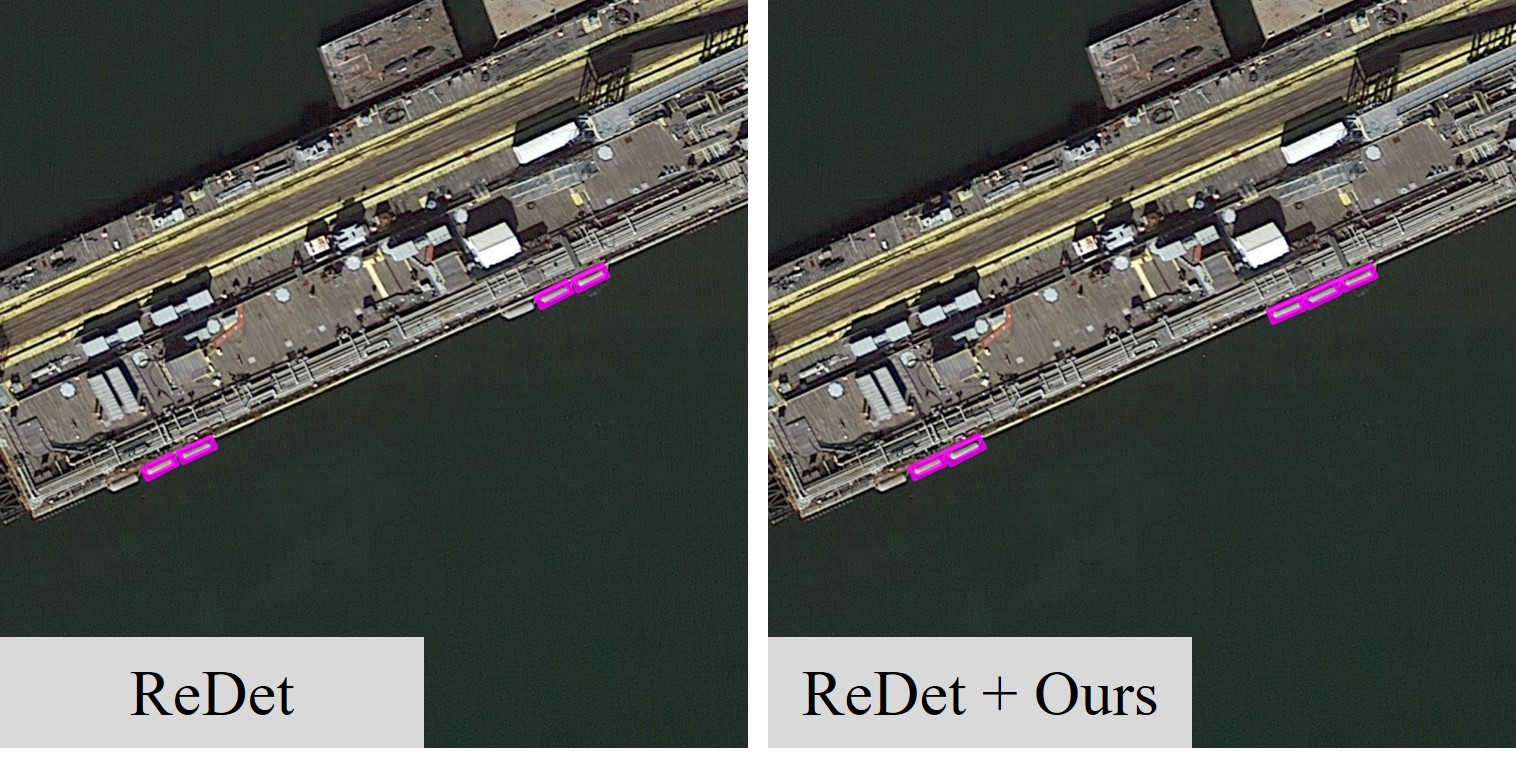}
    \caption{An example of a failure case: The \textcolor{magenta}{\texttt{ship}} detections are all incorrect, but they reinforce each other, leading to an increased number of false ship detections.}
    \label{fig:limitation}
\end{figure}

While our method effectively captures inter-object relationships to improve detection accuracy, there are cases where this approach can lead to undesirable results. Specifically, when a wrong detection occurs for one object, it can propagate errors to nearby objects, particularly if those objects share similar spatial or semantic characteristics. As shown in \cref{fig:limitation}, the false detection of ships in the ReDet model leads to an increased number of false positives for other ship instances when inter-object relationships are captured. This demonstrates that while our method enhances the overall detection performance, incorrect understanding or misidentification of one object may negatively influence the detection of surrounding objects, especially in cases where the objects are spatially or contextually similar.

\section{Conclusion}
\label{limit1}


In this work, we propose a Transformer-based framework to enhance object detection by effectively capturing inter-object and object-background relationships. By integrating the strengths of Transformer models with the cross-modal capabilities of CLIP, our approach not only improves the interaction between Region of Interest (RoI) proposals but also leverages background context for more accurate detections. Extensive experiments on several benchmark datasets demonstrate the effectiveness of our method, yielding consistent improvements over existing detectors. Our analysis further shows that the model reduces scale inconsistency and improves spatial and geometric understanding.

{
    \small
    \bibliographystyle{ieeenat_fullname}
    \bibliography{main}
}


\end{document}